\documentclass[conference]{IEEEtran}
\IEEEoverridecommandlockouts
\usepackage{cite}
\usepackage{amsmath,amssymb,amsfonts}
\usepackage{algorithmic}
\usepackage{graphicx}
\usepackage{amsmath}
\usepackage{textcomp}
\usepackage{xcolor}
\usepackage{amssymb}
\usepackage{algorithm}
\usepackage{algorithmic}
\usepackage{graphicx}
\usepackage{multirow} 
\usepackage{tabularx}
\usepackage{booktabs}
\usepackage{algorithm}
\usepackage{algorithmic}
\usepackage{float}
\usepackage{hyperref}

\def\BibTeX{{\rm B\kern-.05em{\sc i\kern-.025em b}\kern-.08em
            T\kern-.1667em\lower.7ex\hbox{E}\kern-.125emX}}
\begin{document}
\title{MTSA-SNN: A Multi-modal Time Series Analysis Model Based on Spiking Neural Network\\
\thanks{{*} These authors contributed equally to this work.}
\thanks{{\dag} Corresponding Author.}
}


\author{
    \IEEEauthorblockN{Chengzhi Liu$^{a,*,\dag}$, Zheng Tao$^a$, Zihong Luo$^a$, Chenghao Liu$^b$}
    \IEEEauthorblockA{
        $^a$ Xi'an Jiaotong-Liverpool University, \\ $^b$ Shanghai Normal University
    }
    \IEEEauthorblockA{
        \textbf{Email:} \{Chengzhi.Liu21, Zihong.Luo22\}@student.xjtlu.edu.cn
    }
}

\maketitle
\begin{abstract}

Time series analysis and modelling constitute a crucial research area. Traditional artificial neural networks struggle with complex, non-stationary time series data due to high computational complexity, limited ability to capture temporal information, and difficulty in handling event-driven data. To address these challenges, we propose a Multi-modal Time Series Analysis Model Based on Spiking Neural Network (MTSA-SNN). The Pulse Encoder unifies the encoding of temporal images and sequential information in a common pulse-based representation. The Joint Learning Module employs a joint learning function and weight allocation mechanism to fuse information from multi-modal pulse signals complementary. Additionally, we incorporate wavelet transform operations to enhance the model's ability to analyze and evaluate temporal information. Experimental results demonstrate that our method achieved superior performance on three complex time-series tasks. This work provides an effective event-driven approach to overcome the challenges associated with analyzing intricate temporal information. Access to the source code is available at \href{https://github.com/Chenngzz/MTSA-SNN}{https://github.com/Chenngzz/MTSA-SNN}
\end{abstract}


\section{Introduction}

Traditional artificial neural networks (ANNs) have found extensive applications in time series analysis. They serve as a non-parametric, non-linear model capable of effectively capturing complex non-linear relationships within time series data. This is particularly valuable for addressing numerous time series problems since relationships within such data are typically non-linear. Deep neural networks (DNNs), as an extension of ANNs, exhibit a multi-layer structure that automatically learns features and hierarchical information from data. This characteristic enhances the capability of DNNs to analyze complex time series data by capturing patterns at various abstraction levels. For instance, deep learning models like Long Short-Term Memory (LSTM) networks have been widely employed to predict future values or sequences using past time steps\cite{hua2019deep}.ANNs have also been widely applied across a range of applications traditionally addressed by statistical methods, including classification, pattern recognition, prediction, and process control \cite{abiodun2018state}.

However, for complex and volatile time series information, traditional ANNs often face challenges in capturing temporal features accurately. Consequently, Spiking Neural Networks (SNNs), as an alternative approach, have garnered considerable attention. Currently, SNNs have been successfully applied in various time series prediction scenarios, including financial time series forecasting, time series classification \cite{fang2020multivariate}, and real-time online time series prediction \cite{GEORGE202382}.

SNNs rely on discrete signals in continuous time to effectively capture complex time patterns. Nonetheless, current SNN models encounter several challenges. First, the transformation of time series data into a suitable spiking representation poses a significant challenge. Second, the firing times of spiking neurons play a crucial role in model performance, necessitating higher demands for stability and accuracy. Moreover, integrating information from different sources into a single spiking network framework for decision-making involves complex issues related to cross-modal time synchronization and information mapping.

To address these challenges, we propose a Multi-Modal Time Series Analysis model based on Spiking Neural Networks (MTSA-SNN). This model consists of three key components: a single-modal spiking encoder, a spiking joint learning module, and an output layer. The spiking encoder is responsible for transforming time-series information from different modalities into spike signals. It includes alternating layers of feature extraction and neuron layers to selectively process input data from each modality. In the spiking joint learning module, we design a joint learning function and weight allocation mechanism to balance and fuse the complex spike information from multiple modalities. The output layer optimally adjusts the fused spike information to adapt to complex time series analysis tasks.
The main contributions are as follows:

\begin{itemize}

\item A novel Multi-modal Time Series Analysis Model Based on a Spiking Neural Network proposed by us. This model introduces an efficient event-driven approach that overcomes the limitations of traditional time series analysis methods.

\item  We design SNN joint learning functions and a weight allocation mechanism, effectively addressing the balance and fusion of pulsed information.

\item  We synergize wavelet transform with pulse networks to bolster the model's capability in analyzing complex and non-stationary temporal data.

\item Extensive experiments demonstrate the outstanding performance of our approach across multiple complex time series datasets.

\end{itemize}

\section{Related Work}

\subsection{Time Series Forecasting}

Modelling and forecasting time series data is a valuable task in various domains. It has evolved significantly, transitioning from traditional methods to deep learning techniques, resulting in improved prediction accuracy and relevance over time.

Initially, time series forecasting relied on traditional approaches such as the ARIMA model \cite{kong2022time} and Fourier analysis \cite{stein2011fourier}. ARIMA, which includes auto-regressive (AR) and moving average (MA) components with differencing (I) to address non-stationarity, had challenges related to parameter selection and model identification. Fourier analysis was used for frequency domain analysis to identify periodic and seasonal patterns in the data.

Later, deep learning methods such as RNN and LSTM emerged to handle temporal dependencies \cite{10.1007/978-981-19-2456-9_115}. LSTM, an improved version of RNN, performed better with long sequences due to its enhanced memory and forgetting mechanisms, becoming the preferred model for many time series problems. Nonetheless, they encountered challenges related to gradient vanishing and exploding when handling extended sequences, which restricted their practicality.

In contrast to single-modal time series forecasting, multi-modal time series forecasting leverages multiple data sources, such as text, images, and sensor data, to capture a broader perspective, enabling a wider range of pattern and trend recognition. This approach offers benefits like information synthesis, complementarity of different data types, model robustness, and improved generalization. Multi-modal deep learning models use CNN and BiLSTM to extract features from multi-modal time series data. Ensemble models, including probabilistic time series prediction based on Hidden Markov Models \cite{zhang2019high} and stacked ensembles, have been used to enhance accuracy and reduce overfitting.

Specific algorithms, including interpretable ML models and multi-modal meta-learning techniques \cite{chen2021multi}, have been applied in diverse use cases, ranging from early Parkinson's disease detection to time series regression tasks. These applications highlight their potential in various domains, reflecting the diversity and complexity of time series modelling and forecasting. They underscore the evolving methods and technologies that offer robust tools for a broad spectrum of application scenarios.


\begin{figure*}[t]
    \centering
    \vspace{-15pt}
    \includegraphics[width=1.0\textwidth]{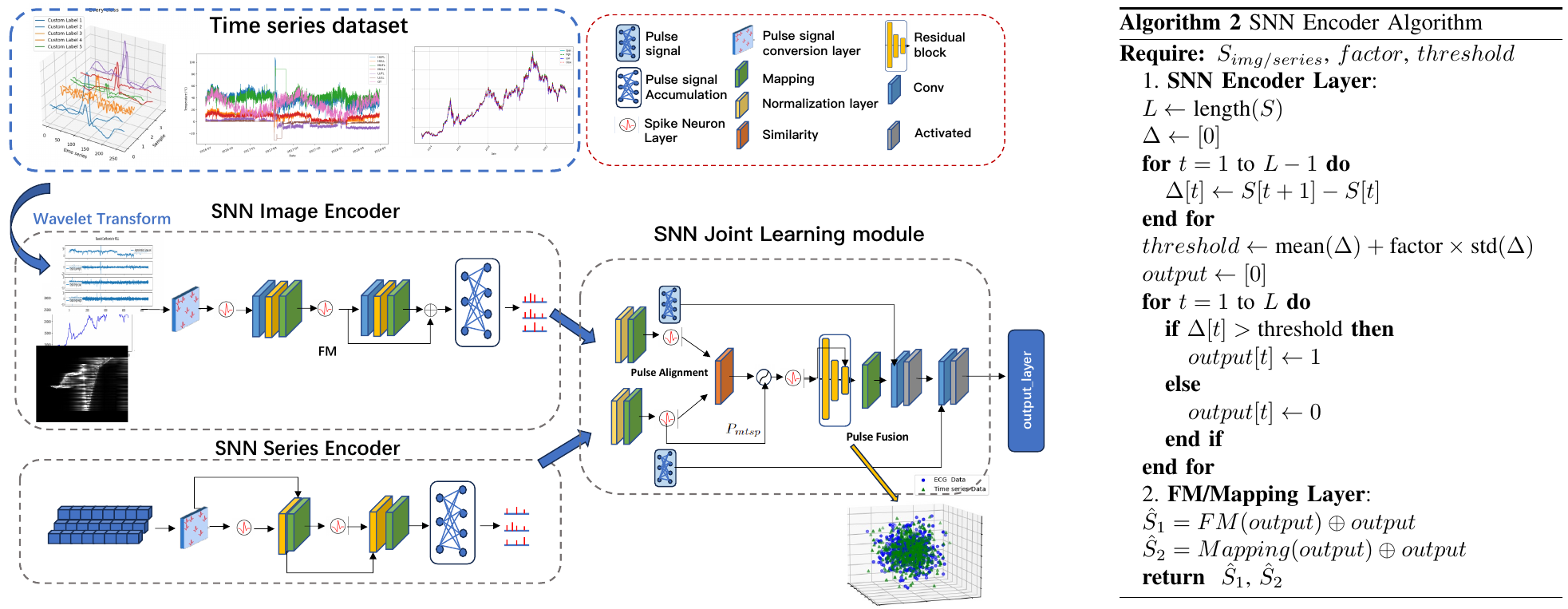}
    \vspace{-10pt}
    \caption{The Overall Structure of MTSA-SNN. SNN Encoder workflow is showed in Algorithm 2.}
    \vspace{-10pt}
    \label{fig:enter-label}
\end{figure*}

\subsection{Spiking Neural Network}

Multi-modal time series models struggle with complex, irregularly and non-uniformly sampled data due to their continuous computations, difficulty in handling event-driven data patterns, and high computational complexity. However, Spiking Neural Networks (SNNs) hold promise in mitigating these challenges. SNNs, a unique class of neural networks that communicate using discrete spike signals in a continuous-time framework \cite{yamazaki2022spiking}, are capable of emulating biological neural systems' sparsity and encoding temporal information \cite{li2020learning}. SNNs find practical application in various time series prediction scenarios, including financial time series forecasting, time series classification \cite{fang2020multivariate}, and real-time online time series prediction \cite{GEORGE202382}.



SNNs pose challenges due to their complex neurons and non-differentiable pulse-based operations. Choosing a multi-modal time series model depends on the problem and data characteristics. The multi-spike network SNN variant is useful for financial time series prediction. Therefore, you should select the most appropriate model based on the problem and data characteristics.

SNNs (Spiking Neural Networks) present challenges due to the complexity of their neurons and the non-differentiable nature of pulse-based operations, making training complex. The choice of a multi-modal time series model should depend on the problem and data characteristics. For instance, a variant like multi-spike networks has proven valuable in time series prediction, especially for non-stationary data \cite{liu2022lstm}. Thus, selecting the right model should align with the problem and data intricacies.

In summary, the proposed MTSA-SNN  model efficiently encodes multimodal information into spikes. It utilizes a spike-based cooperative learning module to effectively map and integrate complex spike information. This method provides an accurate and practical event-driven approach that addresses the analysis of complex and non-stationary temporal information, demonstrating strong performance across multiple time series datasets.

\section{Methodology}
The MTSA-SNN structure consists of three main components: SNN Encoder Module used to extract features from time series data; SNN Joint Learning Module utilizes joint learning and probability distribution methods to map multi-modal signals to a shared joint learning space, enabling the fusion of pulse signals. Output layer used to generate predictions and classification results for multi-modal time series data. The entire workflow is shown in Algorithm \ref{11}.
\begin{algorithm}
	\caption{MTSA-SNN Overall Model}
	\begin{algorithmic}
		\REQUIRE Data of different models
        \STATE \textbf{Input:} $S_{\text{image}}$, $S_{\text{series}}$
        \STATE \textbf{Single-Modal Pulse Encoding Module}:
		\STATE $\hat{S}_1 = \text{Encoder}_i(S_{\text{image}})$, $\hat{S}_1 \in \mathbb{R}^{T \times B \times C \times H \times W}$
		\STATE $\hat{S}_2 = \text{Encoder}_t(S_{\text{series}})$, $\hat{S}_2 \in \mathbb{R}^{T \times B \times C \times T}$  
		\STATE \textbf{SNN Joint Learning Module}:
		\STATE $J_{\text{align}} = \Psi(FT(\hat{S}_1, \hat{S}_2))$, $J_{\text{align}} \in \mathbb{R}^{T \times B \times C \times T}$
		\STATE $J_{\text{fusion}} = \mathbf{JWAM}(\hat{S}_{1/2})$, $J_{\text{fusion}} \in \mathbb{R}^{T \times B \times C \times T}$
        \STATE $J_{MTSA} = \mathbf{Output\_layer}(J_{\text{fusion}}$), $J_{MTSA}  \in  \mathbb{R}^{T \times B \times N}$
		\RETURN $J_{MTSA}$
	\end{algorithmic}
 \label{11}
\end{algorithm}
\subsection{Single-Modal Pulse Encoding Module}
The SNN Image Encoder is a component that processes time-series image information into pulse representations and extracts features. This encoder alternates between the Feature Extraction (FM) module and the Leaky Integrate-and-Fire (LIF) SNN module. Visual information initially passes through the SNN layer to be transformed into a unified and compatible pulse signal format, making it suitable for subsequent network operations. The FM module further performs feature extraction on the visual information converted into pulse signals, including operations such as convolution and pooling. After feature extraction, the pulse signal $\hat{S}_1$ is then passed to the pulse co-learning module.

The SNN Series Encoder is another modality encoder used for pulse-coding and feature extraction of temporal data sequences. These sequence data initially pass through the SNN layer and are then transformed into pulse signals. The network employs alternating operations between mapping layers and neurons. Neurons receive pulse information from the previous layer and membrane potential from the preceding time step in the sequence. By introducing this self-feedback mechanism, the pulse network can utilize membrane potential information from the previous time step to influence the calculations at the current time step. Consequently, the encoder is better equipped to capture the temporal correlations and dynamic changes in time-series data. The pulse information encoded through sequence encoding is $\hat{S}_2$.

Due to the strong temporal information processing capabilities of SNN, we employ the Leaky Integrate-and-Fire (LIF) model to describe the neural dynamics of multi-modal information. The following formula can represent the dynamic equation for the LIF model under continuous-time sequences:
\begin{equation}
\tau_m \frac{dV(t)}{dt} = -(V(t) - V_{\text{rest}}) + R \cdot I(t)
\label{equ:1}
\end{equation}
\begin{equation}
V(t)= V(t-1) +\frac{1}{{\tau}}\left(I(t)-(V(t-1)-V_{\text{rest}})\right)
\label{equ:3}
\end{equation}
$V(t)$ is a membrane potential function concerning time $t$. $V_{\text{rest}}$ represents the resting membrane potential of the neuron. $\tau_m$ is a constant that characterizes the charging and discharging rate of the neuron's membrane potential. 
$I(t)$ is the synaptic pulse input function. $R$ denotes the membrane's responsiveness to input currents. 

When the membrane potential $V(t)$ exceeds the threshold potential $V_{th}$, the neuron is activated and triggers a spike, denoted as $H(t)$.
$\Theta(x)$ is the Heaviside step function, which is 1 when $x \geq 0$ and 0 otherwise. $V_{\text{th}}$ represents the threshold potential. $V_{\text{reset}}$ is the reset potential, to which the membrane potential is reset when the neuron is activated.
\begin{equation}
\begin{cases}
\begin{aligned}
H(t) &= \Theta(V(t) - V_{\text{th}}) \\
V(t) &= V_{\text{reset}}
\end{aligned}
\end{cases}
\label{equ:4}
\end{equation}
A neuron receives multiple pulse signals. Their effects are not independent but accumulate within the neuron, leading to a sustained change in membrane potential. By controlling the pulse frequency and timing, neurons can integrate and encode input information over time. Assuming that N neurons generate multiple pulses at different time points, these pulse timings can be represented by a series of time sequences $\{t_1^{(i)},t_2^{(i)}, ...., t_j^{(i)}\} $. The cumulative effect of multiple pulses can be expressed as $\displaystyle P(t)=  \sum_{i=1}^N\sum_{j=1}^{j}f(t-t_j^{(i)})$.

$f(t)$ represents the Dirac Delta function, signifying the generation of a pulse at the firing time. $P(t)$ is the output of the cumulative effect of multiple pulses, which corresponds to the pulse output of the encoder $\hat{S}_1$ \& $\hat{S}_2$. Algorithm 2 is the workflow of the SNN encoder. 

\subsection{Multi-Modal pulse Joint Learning Module}
The pulse signals extracted from different encoders are first subjected to normalization and mapping operations before input into a unified pulse co-learning module. The pulse signals $\hat{S}_1$ and $\hat{S}_2$ obtained from two heterogeneous spaces are then transformed from the time domain to the frequency domain through Fourier transformation. Fourier transformation $FT(s)$ decomposes the signal into different frequency components, which aids in analyzing the frequency domain characteristics of different modal signals.
\begin{align}
    \hat{S}_1 & = \text{Encoder}_i(S_{\text{image}}) \in \mathbb{R}^{N \times D_i} \\
    \hat{S}_2 & = \text{Encoder}_t(S_{\text{series}}) \in \mathbb{R}^{N \times D_t} 
    \label{equ:123}
\end{align}
To better integrate and align the information from two different signal spaces, we introduce a joint learning function denoted as $\Psi$. This function aims to adjust the feature representations of the signals, mapping the signals from space $D_i$ and space $D_t$ to a common frequency domain space. During the training process, this function is continuously adjusted to make the pulse information in different modalities more consistent, achieving effective fusion and alignment of heterogeneous signals. $J_{align}$ denotes the fusion of pulse information in the joint learning space. $D_j$ is the dimension of joint learning space, where data from different modalities coexist in a shared representation. 
\begin{align}
    FT(s) & = \int_{-\infty}^{\infty} \hat{S} \odot e^{-iwT} \in \mathbb{R}^{N \times D_j}\\
    J_{align} & = \Psi \biggl(FT(\hat{S}_{1}), FT(\hat{S}_{2})\biggr) 
    \label{equ:4567}
\end{align}
We introduce a more effective pulse-based joint weight allocation mechanism (JWAM). This mechanism involves mapping the similarity results in $sim$ of multi-modal pulse signals into different spatial dimensions of the probability distribution matrix ($P_{mtsa}$). The similarity probability distribution is adaptively adjusted based on the features of each modality and their relative importance to achieve information fusion. $P_{mtsa}$ integrates information from various modalities, providing a quantitative method for scoring cross-modal information representation. $sim$ is a metric function used to measure the similarity between two pulse information representations in heterogeneous spaces. This function employs the Euclidean distance calculation method to assess the similarity between different modality representations. $\sigma^2$ is used to adjust the sensitivity of the similarity measurement function. It is worth noting that it can dynamically adapt based on the distribution information of different modality features, enhancing the robustness and adaptability of similarity measurements.
\begin{align}
    sim & = \exp\left(-\frac{|\sum_{dim=1}^{i,t}(\hat{S}_1- \hat{S}_2)|}{2\sigma^2}\right) \\
    P_{mtsa} & = \frac{\exp\bigl(sim_i, sim_t \bigr)}{\sum_{dim=1}^j \exp\left(Sim_i, Sim_t \right)}
    \label{equ:8}
\end{align}
Furthermore, matrix transformations of the information in the joint space are utilized to adjust pulse signals.
This operation aims to optimize the feature space while taking into consideration information from different modalities in order to better accommodate the characteristics of pulse sequences from other modalities. Additionally, we interact this process with cross-modal probability distributions to obtain the pulse fusion representation denoted as $J_{fusion}$. This can be expressed as: 
\begin{equation}
   J_{fusion} = Softmax\left(\frac{\hat{S}_{1/2} \odot J_{align}}{\sqrt{D_j}}\right)\odot P_{mtsa} 
\end{equation}

The Output layer is responsible for two major tasks: predicting and classifying information from multi-modal time series pulse fusion data. It employs network layer techniques such as residual connections and ReLU to transform the fused information into a common format, making it available for various downstream tasks. 

\subsection{Pulse Signal Processing Based on Wavelet Transform}
To effectively address the non-stationary, non-linear characteristics and constraints in multi-scale feature analysis of time-series data, we employ the wavelet transform analysis method. Wavelet transform possesses exceptional time-frequency locality and multi-scale analysis capabilities, making it more suitable for capturing local features of signals at different time and frequency scales. The MTSA-SNN network based on wavelet transform can capture richer feature representations, endowing it with a significant advantage in handling non-stationary signals, extracting critical signal features, and analyzing signals across multiple scales.

\begin{figure}[H]
    \centering
    \vspace{-10pt}
\includegraphics[width=0.481\textwidth]{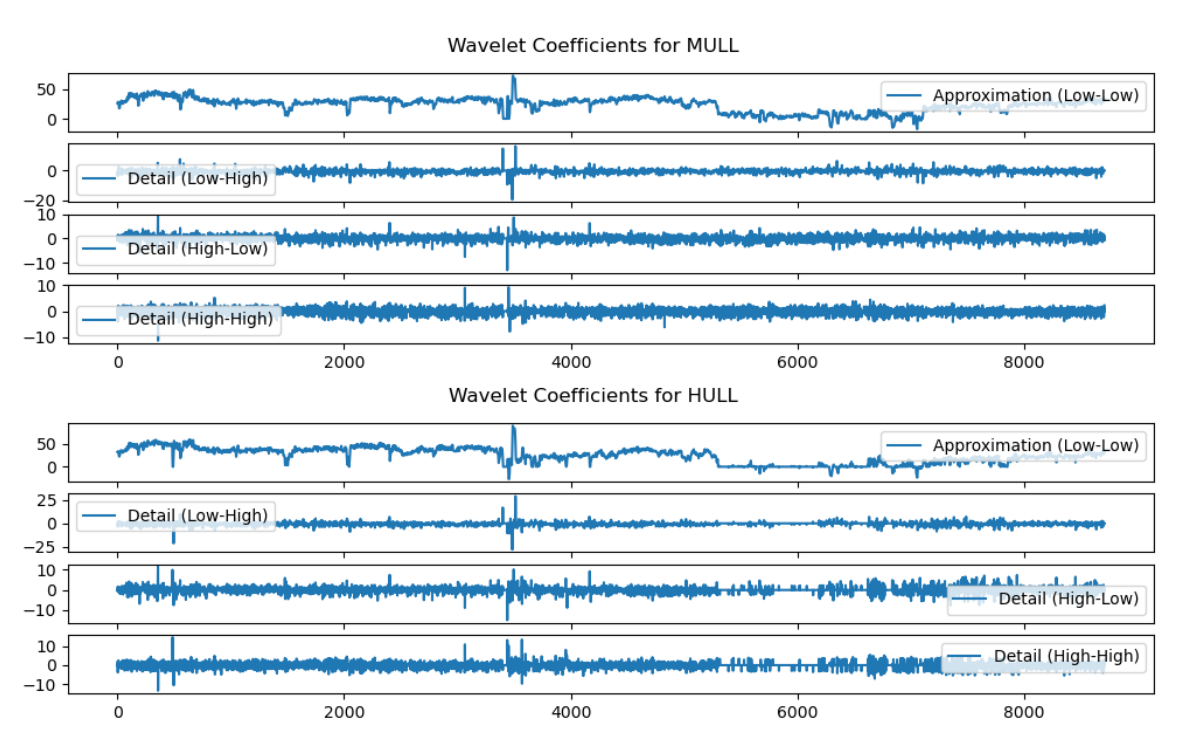}
    \vspace{-10pt}
    \caption{ETT dataset signal features across different frequency and spatial scales. (LL captures low-frequency signal components. LH and HH capture high-frequency components in both low and high-frequency signals. HL contains low-frequency components of high-frequency signals.)}
    \label{fig:stock_market}
\end{figure}

\begin{figure}[H]
    \centering
    \vspace{-20pt}
    \includegraphics[width=0.5\textwidth]{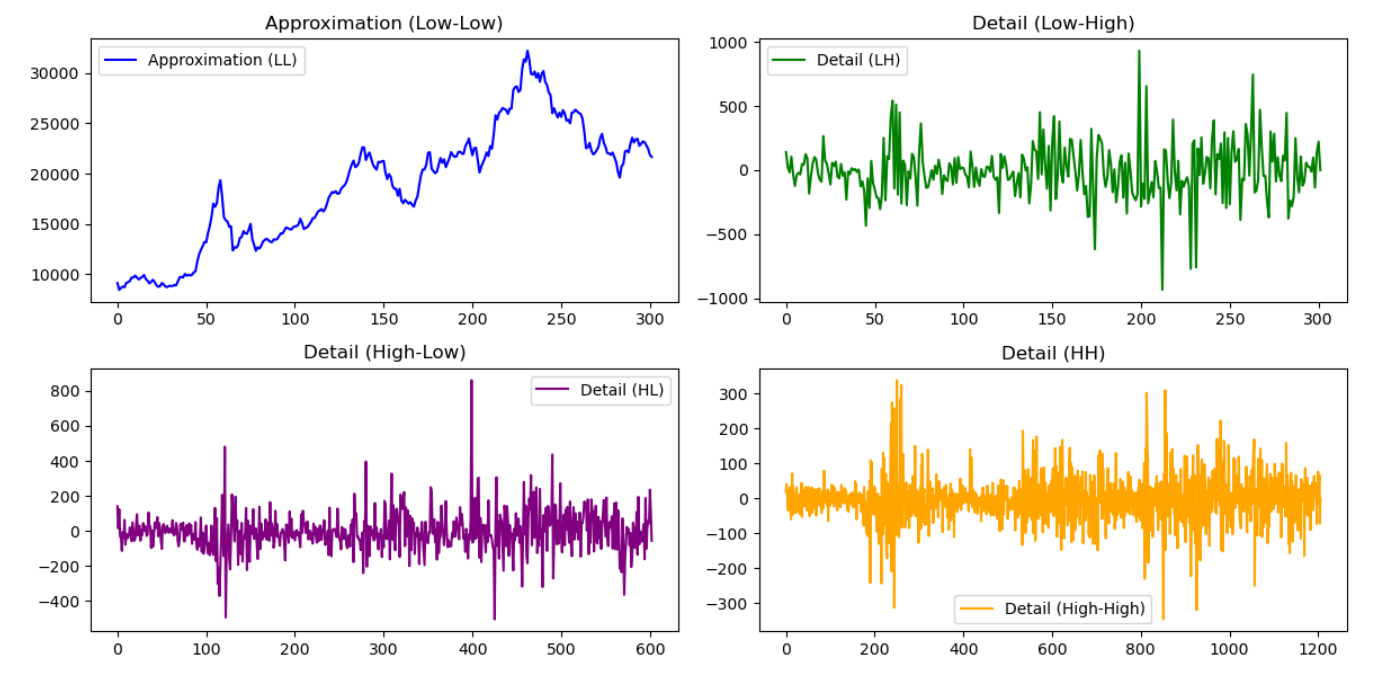}
    \vspace{-10pt}
    \caption{Stock prediction dataset signal features across different frequency and spatial scales}
    \label{fig:trans_vis}
    \vspace{-10pt}
\end{figure}

MTSA-SNN employs wavelet transform to decompose input signals into four subbands: LL, LH, HH and HL, which represent distinct signal characteristics in terms of different frequencies and spatial scales.  This multi-scale and multi-frequency analysis approach equips the MTSA-SNN model with a comprehensive understanding of multimodal data, enhancing its learning capabilities. As illustrated in Fig. \ref{fig:stock_market} and Fig. \ref{fig:trans_vis}, the temporal visualizations of these four subbands in the ETT and stock prediction datasets demonstrate the effectiveness of this multi-scale analysis.
\begin{figure}[H]
    \centering
    \includegraphics[width=0.43\textwidth]{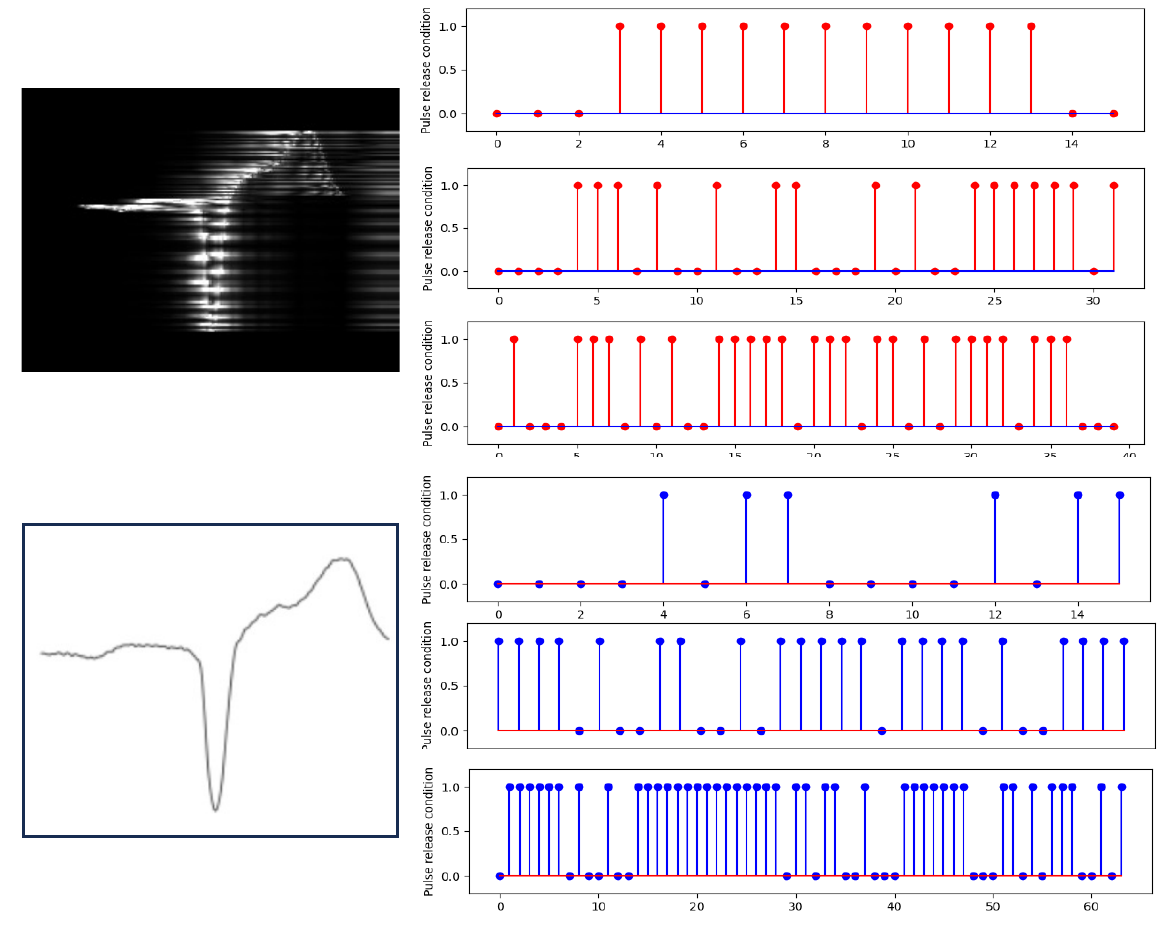}
    \vspace{-10pt}
    \caption{Data (wavelet transform) converted into pulse signals by MTSA-SNN (above) \& Original data converted into pulse signals by MTSA-SNN}
    \label{fig:wave2}
    \vspace{-10pt}
\end{figure}
Fig. \ref{fig:wave2} depicts the pulse network outputs based on the MIT-BIH dataset with different processing methods. It is evident that the pulse output subjected to wavelet transform more accurately captures the features of multimodal signals, resulting in a more stable and effective neural activation.

\section{EXPERIMENT}
\subsection{Datasets}
We conduct experimental evaluations for classification and regression tasks on two traditional time series datasets, MIT-BIH Arrhythmia (MIT-BIH)\cite{moody2001impact} and Electricity Transformer Temperature (ETT)\cite{zhou2021informer}. Additionally, we perform a market forecasting analysis on relevant stock indices of the Chinese stock market from June 6, 2013, to June 6, 2023, covering a ten-year period, focusing on the opening and closing prices.

\subsection{Comparison with other methods}
MTSA-SNN demonstrates remarkable performance advantages in the field of biological time-series data analysis. The experimental results in Table \ref{tab:comparison} demonstrates that our model has achieved advanced performance in the detection of cardiac arrhythmias in multimodal electrocardiogram data. With a dataset classification accuracy of 98.75\%, MTSA-SNN markedly outperforms previous leading algorithms. This is attributable to the effective simulation of the neural signal conduction process in biological systems through MTSA-SNN's pulse-based fusion approach, resulting in significant performance advantages.
\begin{table}[!ht]
    \caption{Comparison of MTSA-SNN with other methods on the five-class MIT-BIH dataset}
    \label{tab:comparison}
    \vspace{-10pt}
    \centering
    \vspace{10pt} 
    \begin{tabular}{lccc}
        \toprule
          Network & Accuracy (\%)$\uparrow$ & F1(\%) $\uparrow$&  Precision(\%)$\uparrow$\\
        \midrule
         Mousavi et al.\cite{liu2022lstm}     & 97.62 & 85.82 &  91.46 \\
          Yang et al.\cite{YANG201822}    & 97.76  &  88.28 &  94.34  \\
          Hammad et al.\cite{hammad2020multitier} & 98.00 & 89.70 & 86.55 \\
          Xing et al.\cite{xing2022accurate} & 98.26 & 89.09 & - \\
          MTSA-SNN (ours) & \textbf{98.75} & \textbf{94.31} & \textbf{94.62}\\
        \bottomrule
    \end{tabular}
    \vspace{-8pt}
\end{table}
In addition, our method exhibits outstanding performance in various prediction tasks, including transformer temperature monitoring and stock market forecasting. Analyzing the results presented in Table \ref{tab:ETT}, our model demonstrates the lowest MAE and MSE across four different time steps in the ETT dataset. Furthermore, in Table \ref{tab:3}, MTSA-SNN achieves remarkably low errors of 0.96 and 1.15 in the stock market price prediction task compared to traditional time-series prediction models such as LSTM and XGBoost. MTSA-SNN, by converting complex and diverse multimodal time series data into a pulse-based representation, significantly enhances the model's predictive and analytical capabilities regarding time-series information.
\vspace{-9pt}
\begin{table}[ht]
\caption{Comparison of MTSA-SNN with other methods on ETT dataset}
\label{tab:ETT}
\resizebox{0.48\textwidth}{!}{
\begin{tabular}{cc|cccccccccccc}
\hline
\multicolumn{2}{c|}{Methods} & \multicolumn{2}{c|}{NLinear \cite{zeng2023transformers}} & \multicolumn{2}{c|}{DLinear \cite{zeng2023transformers}} & \multicolumn{2}{c|}{Autoformer\cite{wu2021autoformer}} & \multicolumn{2}{c|}{Informer\cite{zhou2021informer}}& \multicolumn{2}{c|}{MTSA-SNN (ours)}   \\ \hline
\multicolumn{2}{c|}{Metric} & MSE$\downarrow$ & MAE $\downarrow$ & MSE & MAE & MSE & MAE & MSE & MAE & MSE & MAE \\\hline
\multicolumn{1}{c|}{\multirow{4}{*}{\rotatebox{90}{$ETT$}}} & 96  & 0.374 & 0.394 & 0.375  & 0.399 & 0.449 & 0.459& 0.865 & 0.713 & \textbf{0.235} & \textbf{0.247}\\
\multicolumn{1}{c|}{} & 192 & 0.408 & 0.415 &  0.405 & 0.416 & 0.500& 0.482& 1.008 & 0.792 & \textbf{0.345} & \textbf{0.371} \\
\multicolumn{1}{c|}{} & 336 & 0.429 & 0.427 & 0.439 &  0.443 & 0.521&  0.496& 1.107 & 0.809 & \textbf{0.358} & \textbf{0.362}  \\
\multicolumn{1}{c|}{} & 720 & 0.440 & 0.453 & 0.472 &  0.490 & 0.514& 0.512   & 1.181 & 0.865 & \textbf{0.396} & \textbf{0.439}  \\ 
\hline
\end{tabular}}
\end{table}
\vspace{-15pt}
\begin{table}[ht]
    \caption{Comparison of MTSA-SNN with other methods on Stock market price prediction dataset}
    \label{tab:3}
    \centering
    \vspace{-5pt} 
    \begin{tabular}{lcccc}
        \toprule
          Network & LSTM& XGBoost &  LSTM-XGBoost& MTSA-SNN (ours) \\
        \midrule
         MAE $\downarrow$     & 2.465 & 2.317 & 1.394 & \textbf{0.961} \\
         MSE $\downarrow$    & 2.839 & 2.285 &  1.461& \textbf{1.152} \\     
        \bottomrule
    \end{tabular}
    \vspace{-10pt}
\end{table}

\begin{figure}[ht]
    \centering
    \includegraphics[width=0.45\textwidth, height= 6.4cm]{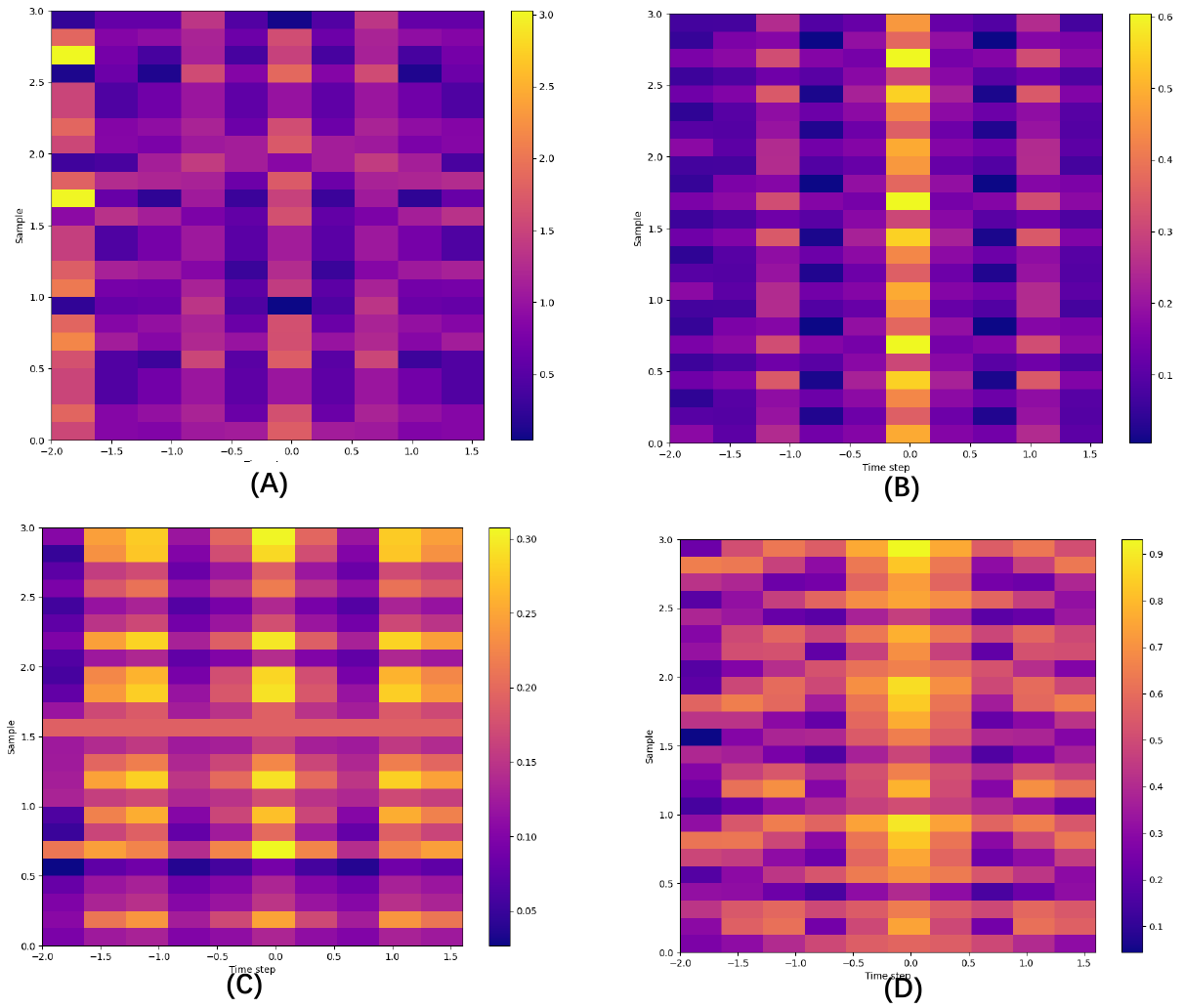}
    \caption{The heatmap of MTSA-SNN's various component neuron activations. Specifically, (A) and (B) represent the neuron activation patterns after the time series information passes through the image encoder and sequence encoder of MTSA-SNN. (C) demonstrates the fused output after the joint learning process for the original temporal information.  (D) represents the pulse fusion after applying wavelet transform in MTSA-SNN.}
    \label{fig:exp_result2}
    \vspace{-5pt}
\end{figure}

\subsection{Ablation Study}
We conduct a comprehensive ablation study to evaluate different components of the MTSA-SNN model. As shown in Fig. \ref{fig:exp_result2}, we present pulse signal output heatmaps for different components at the same time step using the MIT-BIH dataset. The brightness of the colours in the figure represents the activation levels of neurons. 
In comparison to the activation patterns from single-modal encoders, the joint learning module of MTSA-SNN activates more neurons, thus enriching the representation of temporal information. Furthermore, the application of wavelet transform enhances the representation of temporal information within the MTSA-SNN. This suggests that joint learning of pulses effectively balances multi-modal pulse signals and fuses them together. Simultaneously, wavelet transformation contributes to enhancing the representation of temporal information in the pulse network.

\begin{figure}[!ht]
    \centering
    \vspace{-10pt}
    \includegraphics[width=0.44\textwidth, height = 10cm]{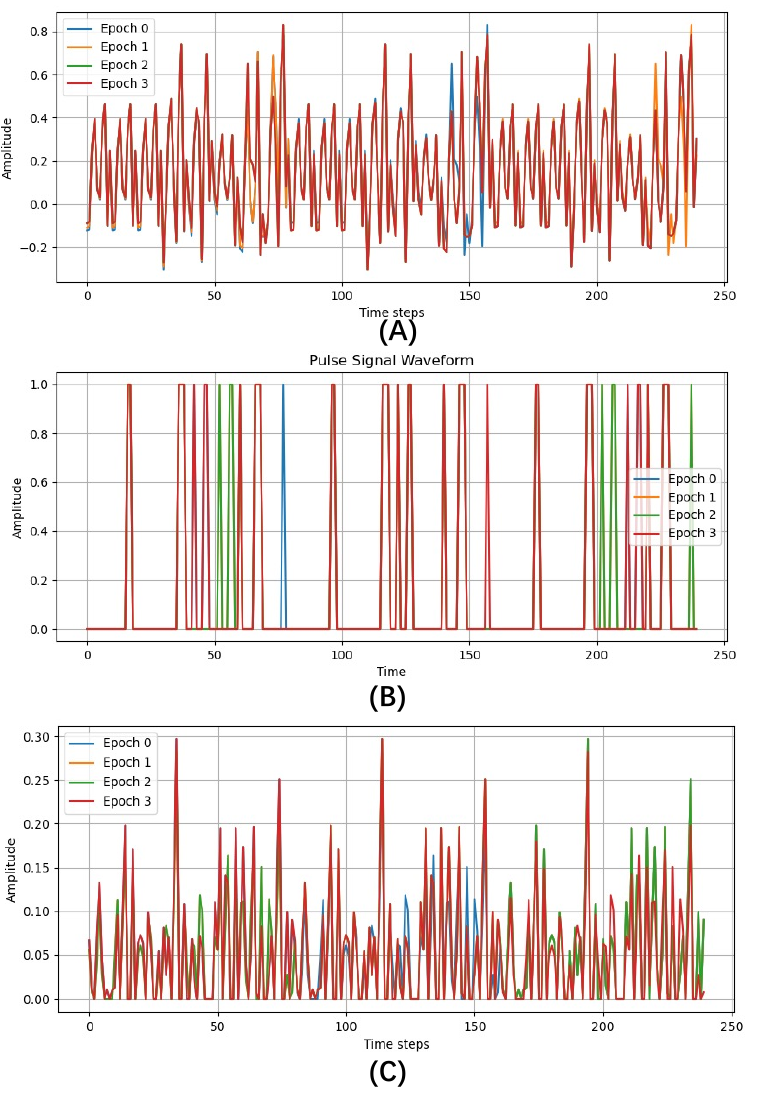}
    \vspace{-10pt}
    \caption{The spectral analysis of waveform plots during training(The first four epochs). (A) and (B) show the waveforms generated by the single-modal encoder, revealing unstable characteristics of the pulse signals and relatively weak robustness in the individual modality. (C) shows the output of the MTSA-SNN model, exhibiting significant frequency domain stability as it consistently remains within a defined range of amplitudes.}
    \label{fig:Exp_result1}
    \vspace{-8pt}
\end{figure}

In addition, we analyze the spectral information of the waveform plots during the training process of the single-modal encoder and the joint learning module. In Fig. \ref{fig:Exp_result1}, the horizontal axis represents the time steps, while the vertical axis represents the amplitude. This indicates that the MTSA-SNN model effectively integrates and analyzes multi-modal signals while enhancing the overall robustness of the model.

\section{Conclusion}

In this paper, we introduce an innovative Multi-modal Time Series Analysis Model based on the Spiking Neural Network. The model's pulse encoder is designed to uniformly pulse-code multi-modal information. The pulse joint learning module is employed to effectively integrate complex pulse-encoded data. Additionally, we incorporate wavelet transform operations to enhance the model's capability to analyze and evaluate time series data. Experimental results on three distinct time series datasets demonstrate the outstanding performance of our proposed approach across multiple tasks.
\bibliographystyle{elsarticle-num} 
\bibliography{conference}

\end{document}